\documentclass{article}

\usepackage{arxiv}

\usepackage[utf8]{inputenc} 
\usepackage[T1]{fontenc}    
\usepackage{hyperref}       
\usepackage{url}            
\usepackage{booktabs}       
\usepackage{amsfonts}       
\usepackage{nicefrac}       
\usepackage{microtype}      
\usepackage{lipsum}
\usepackage{graphicx}
\graphicspath{ {./images/} }

\title{Explainable AI (XAI) for Arrhythmia detection from electrocardiograms}

\author{
 Joschka Beck \\
  Biomedical Signals and Systems Group\\
  University of Twente\\
  Drienerlolaan 5, Enschede 7522 NB\\
  The Netherlands \\
   \And
 Arlene John \\
  Biomedical Signals and Systems Group\\
  University of Twente\\
  Drienerlolaan 5, Enschede 7522 NB\\
  The Netherlands\\
  \texttt{a.john@utwente.nl} \\
}

\begin{document}
\maketitle
\begin{abstract}
Advancements in deep learning have enabled highly accurate arrhythmia detection from electrocardiogram (ECG) signals, but limited interpretability remains a barrier to clinical adoption. This study investigates the application of Explainable AI (XAI) techniques specifically adapted for time-series ECG analysis. Using the MIT-BIH arrhythmia dataset, a convolutional neural network-based model was developed for arrhythmia classification, with R-peak–based segmentation via the Pan–Tompkins algorithm. To increase the dataset size and to reduce class imbalance, an additional 12-lead ECG dataset was incorporated. A user needs assessment was carried out to identify what kind of explanation would be preferred by medical professionals. Medical professionals indicated a preference for saliency map–based explanations over counterfactual visualisations, citing clearer correspondence with ECG interpretation workflows. Four SHapley Additive exPlanations (SHAP)-based approaches—permutation importance, KernelSHAP, gradient-based methods, and Deep Learning Important FeaTures (DeepLIFT)—were implemented and compared. The model achieved 98.3\% validation accuracy on MIT-BIH but showed performance degradation on the combined dataset, underscoring dataset variability challenges. Permutation importance and KernelSHAP produced cluttered visual outputs, while gradient-based and DeepLIFT methods highlighted waveform regions consistent with clinical reasoning, but with variability across samples. Findings emphasize the need for domain-specific XAI adaptations in ECG analysis and highlight saliency mapping as a more clinically intuitive approach.
\end{abstract}


\section{Introduction} 
\label{sec:introduction}
Cardiovascular diseases (CVDs) are the leading cause of death worldwide \cite{Joseph_2025}. In the European Union alone there were 1.71 million deaths in 2021, that made up 32.4\% of all premature deaths \cite{european_union}. These deaths cost the European Union an estimated € 282 billion annually. Of these costs, 55\% is from direct health care and long-term care, productivity loss and informal care make up the rest \cite{Luengo-Fernandez_2023}. An approach to reducing the CVD-related deaths and associated costs is through early detection \cite{Saeed_2011}. This can be achieved by detecting and identifying arrhythmias- abnormalities in the heartbeats- from the electrocardiogram (ECG) signals. In hospitals, short ECG recordings are often captured, and while arrhythmias may sometimes appear in these readings, they can also be missed. Moreover, symptoms often develop gradually, so long-term recordings are generally more effective, as they increase the likelihood of detecting any arrhythmia.\\
With recent advancements in wearable technology, high-risk individuals can record their own electrocardiograms (ECGs) with increasing ease. However, analyzing ECG recordings for abnormalities is a meticulous process that requires the expertise of trained professionals. The widespread availability of wearable devices, combined with the benefits of long-term monitoring, has led to a substantial increase in the volume of ECG data, making manual review by medical practitioners impractical. Recent developments in machine learning (ML) and deep learning (DL) have demonstrated strong capabilities in processing large datasets rapidly and, when properly implemented, achieving high diagnostic accuracy. In recent years, multiple studies have shown that these techniques are highly effective for the detection of arrhythmias from ECG signals \cite{Ansari_2023, Boulif_2023}. These models can greatly accelerate the diagnostic process, with most achieving accuracies above 95\%. However, it remains essential that the final judgment is made by a physician \cite{Alamatsaz_2024}. Although these models demonstrate high accuracy, they typically provide little information beyond the detected arrhythmia. While they can classify the type of arrhythmia and output a corresponding confidence score, this is often the extent of the available information \cite{Ahsan_2019, Singh_2022}. Unlike human experts, such models do not offer reasoning or justification for their predictions. This lack of interpretability is problematic, as medical practitioners are required to rely on what effectively functions as a “black box” system. Enhanced interpretability would benefit not only trained specialists but also healthcare providers in settings where cardiology expertise is unavailable, such as in certain low-resource regions \cite{who}. The challenge of limited interpretability is not unique to ECG analysis; it is a broader issue affecting many modern deep-learning models. This concern has led to the emergence of the field of eXplainable Artificial Intelligence (XAI), which focuses on improving the transparency of advanced AI systems. However, most existing XAI techniques have been developed for image or tabular data and are not directly suited to time-series analysis, thereby limiting their applicability to ECG interpretation \cite{Schlegel_2019}. This study evaluates strategies to enhance the explainability of current deep learning–based arrhythmia detection models, leading to the contributions of this work:
\begin{itemize}
    \item A qualitative analysis into what healthcare practioners would prefer with regard to explainability of automated arrhythmia detection models.
    \item An analysis of XAI techniques that can aid in the interpretation of automated arrhythmia detection models applied on time-series ECG signals.

\end{itemize}
The remainder of this article is structured as follows: Section \ref{rel_lit} reviews the related literature, Section \ref{sec:methodology} outlines the methodology, and Section \ref{sec_results} presents the results. The discussion is provided in Section \ref{sec_discussions}, followed by the conclusions in Section \ref{sec_conclusions}.
\section{Related literature}
\label{rel_lit}
\subsection{Machine learning for arrhythmia detection from electrocardiogram}
In recent years, numerous machine learning (ML) and deep learning (DL) approaches have been applied to improve arrhythmia detection in electrocardiogram (ECG) signals. State-of-the-art models have achieved very high performance, with some reporting accuracies, sensitivities, and specificities exceeding 98\% \cite{Ansari_2023}. Nonetheless, further improvements are possible, particularly in addressing dataset imbalances and ensuring well-defined evaluation protocols \cite{Patra_2023}. Due to these imbalances, most existing algorithms—despite the wide variety of arrhythmia types—are designed to detect six or fewer classes. Developing a high-performing model therefore requires careful consideration of available techniques and their suitability for the task.

Various ML and DL methods have been explored for arrhythmia detection; in this work, we focus on DL techniques, which have demonstrated superior performance \cite{Ansari_2023}. Convolutional Neural Networks (CNNs) are employed for their ability to capture spatial hierarchies and patterns using small, trainable filters (kernels). These kernels extract local information from ECG signals, such as heartbeat shape and duration. CNNs can operate in both one-dimensional (time-series) and two-dimensional (image) formats. However, their fixed input size limits the temporal range of analysis, potentially overlooking arrhythmias that manifest over extended periods.

To address this limitation, Long Short-Term Memory (LSTM) networks, an extension of Recurrent Neural Networks (RNNs), are utilized for their capability to retain information over longer time spans and model long-term temporal dependencies, making them well-suited for event detection in time-series data \cite{Alamatsaz_2024}. Nevertheless, certain explainability techniques perform suboptimally with recurrent architectures, posing challenges for model interpretability.
\subsection{Explainable Artificial Intelligence (XAI)}
Explainable Artificial Intelligence (XAI) encompasses a set of methods and tools designed to make the decision-making processes of AI models understandable to humans. It can be likened to the reasoning process a human might use to justify a decision, providing insight into why a model produces a particular output \cite{Rudresh_2023}.

The understandability of an AI model can be improved in two ways: interpretability and explainability \cite{Ali_2023}. Although these terms are often used interchangeably, interpretability generally refers to how easily the internal mechanisms of a model can be comprehended. For instance, simple models such as linear regression or decision trees are inherently more interpretable than complex deep learning (DL) architectures. Explainability, on the other hand, focuses on how well a human can understand the reasoning behind a model’s decision, particularly when dealing with complex models that require additional tools or methods to clarify their outputs.

XAI techniques can be applied at two levels: intrinsic and post-hoc \cite{Yang_2019}. Intrinsic methods incorporate explainability directly into the model’s architecture, producing both the prediction and an accompanying rationale. Post-hoc methods, in contrast, generate explanations after model training, using the original model to extract interpretive insights. Furthermore, XAI can operate on a global or local scale. Global explanations describe the overall behavior of a model, whereas local explanations focus on the reasoning behind a single prediction \cite{Yang_2019}.

While numerous global-scale XAI methods exist, most are not well-suited for time-series data such as ECG signals. This limitation stems from two main factors. First, the majority of AI applications, and consequently, XAI development, are concentrated on image, text, or feature-based inputs \cite{Schlegel_2019, Liang_2021}. Second, explaining time-series models is inherently more challenging, as predictions depend not only on individual time points but also on past and future observations and their temporal relationships \cite{Katy_2024}. Even when XAI techniques can be adapted to time series, they often lack reliability by failing to capture these dependencies. In this work, we employ several XAI methods capable of operating on time-series data, as detailed below:
\begin{itemize}
    \item Permutation importance: This method assesses feature relevance by permuting (i.e., shuffling) data points and measuring the resulting change in model error. While it can be applied to time-series data, random shuffling may disrupt temporal dependencies, leading to unreliable explanations \cite{Singh_2022}.
    \item CAM and grad-CAM: The Class Activation Map (CAM) technique highlights important regions in the input using the learned weights from training. Grad-CAM extends this approach by leveraging the gradients of the target class, making it more flexible. An additional variant, k-Grad-CAM, emphasizes the most salient features, enabling more focused visualizations. Singh \textit{et al.} \cite{Singh_2022} identified k-Grad-CAM as the most effective method for ECG signal analysis.
    \item SHAP: SHapley Additive exPlanations (SHAP) is derived from Shapley values in game theory. For time-series data, it assigns an importance score to each data point by quantifying its contribution to the final decision across all possible feature combinations. However, the method’s reliance on random perturbations can result in unstable explanations \cite{Singh_2022}.
    \item DeepLIFT: Deep Learning Important FeaTures (DeepLIFT) compares the activation of each neuron to a predefined reference activation and assigns contribution scores based on the difference \cite{Shrikumar_2019}. Unlike gradient-based methods, it is not affected by zero or discontinuous gradients. However, it requires a custom activation function, necessitating either model retraining or the use of a surrogate model.
\end{itemize}
In most cases, DeepLift and SHAP are found to be the most robust methods when trying to explain time series models. DeepLIFT is concluded to have the best quality for CNN models, and SHAP being more robust to adjust to all types of deep-learning models \cite{Schlegel_2019}. With regard to arrhythmia detection from ECG signals, SHAP-based explanations are commonly used. It can be applied in multiple ways. It can be used on the raw signal \cite{Alamatsaz_2024} or on features extracted from ECG signals \cite{Sathi_2024}. 
\section{Methodology}
\label{sec:methodology}
\subsection{Dataset}
When selecting databases for arrhythmia classification in ECG signals, the MIT-BIH Arrhythmia Database remains one of the most widely used and comprehensive resources \cite{Boulif_2023}. It consists of 48 half-hour recordings of two-lead ECG signals, with each heartbeat annotated by cardiologists according to the arrhythmia type \cite{Moody_2001, Goldberger_2000}. To enhance model robustness and generalization, it is recommended to train and evaluate using different datasets \cite{Boulif_2023}. Accordingly, a second dataset was incorporated in this study, collected by Zheng et al. \cite{Zheng_2022, Zheng_2020}, which contains 12-lead ECG recordings, in contrast to the two-lead format of MIT-BIH.
\subsection{Pre-processing}
The raw data obtained from the database requires preprocessing to ensure it is suitable for input into an AI model. Proper preprocessing improves data quality, reduces noise, and ensures consistency across samples, which is essential for reliable model training and evaluation. In this study, the following steps were implemented to prepare the data for analysis:
\begin{enumerate}
    \item Segmentation and filtering: For AI-based analysis, it is practical to treat each individual heartbeat as a separate sample, as each beat is associated with a specific class label. ECG recordings, however, often contain low- and high-frequency noise arising from sources such as power line interference, baseline wander, and electromyographic activity \cite{Alamatsaz_2024}. To address both segmentation and noise removal, the Pan-Tompkins algorithm—a widely used method for QRS-complex detection—was employed \cite{Pan_1985}. This algorithm filters the ECG signal using frequency-based filters, which are then used as the input to the models. Peak detection on the filtered signal is carried out through a three-phase procedure. Once the R-peaks are identified, the signal can be segmented into individual heartbeats, with 200 ms preceding and 400 ms following each peak, resulting in 0.6-second segments per sample. This segment length falls within an optimal range for model performance, as segments shorter than 0.4 seconds can negatively impact accuracy \cite{Chen_2024}. Following segmentation, the dataset comprised 112,575 samples, with the majority corresponding to normal sinus rhythm (approximately 70,000 samples). Other arrhythmia classes contained fewer than 10,000 samples each, with rare arrhythmias represented by even smaller counts.
\item Handling class imbalance: To address class imbalance, classes with fewer than 2,000 samples were excluded from the analysis. The remaining classes used for model development included Paced Beat, Atrial Premature Beat, Left Bundle Branch Block, Right Bundle Branch Block, Premature Ventricular Contraction, and Normal Rhythm. Samples from these classes were supplemented with data from a second dataset \cite{Zheng_2022}, integrated with the MIT-BIH dataset using the corresponding SNOMED CT codes. The combined dataset was then divided into training and validation sets with and an 80/20 split.
\end{enumerate}
\subsection{Model Architecture}
The model architecture was initially based on the design presented in \cite{Alamatsaz_2024}, which demonstrated strong performance using a simple Long Short-Term Memory (LSTM)-based structure. However, LSTM layers pose challenges for explainability in the context of the XAI methods evaluated in this study. Consequently, the original architecture was modified by removing the LSTM layers and retaining only convolutional layers. The resulting architecture is summarized in Table \ref{model_cnn}. During training, a validation-based callback was employed to save the model achieving the best performance.

\begin{table}[]
\caption{Structure of the CNN-based model for arrhythmia detection}
\label{model_cnn}
\centering
\begin{tabular}{|c|c|c|}
\hline
\textbf{Layer (type)} & \textbf{Output Shape} & \textbf{Number of parameters} \\ \hline \hline
Input & 216x1 & 0 \\ \hline
Conv1D & 167x64 & 3,264  \\ \hline
MaxPooling1D & 83x64 & 0 \\ \hline
Dropout & 83x64 & 0 \\ \hline
Conv1D & 74x32 & 20,512  \\ \hline
MaxPooling1D & 37x32 & 0 \\ \hline
Dropout & 37x32 & 0 \\ \hline
Conv1D & 33x16 & 2,576 \\ \hline
MaxPooling1D & 16x16 & 0 \\ \hline
Dropout & 16x16 & 0 \\ \hline
Flatten & 256 & 0 \\ \hline
Dense & 32 & 8,224  \\ \hline
Dropout & 32 & 0 \\ \hline
Dense & 16 & 528 \\ \hline
Dense & 23 & 391 \\ \hline
\end{tabular}
\end{table}
\subsection{Opinion of Medical Professionals}
\begin{figure}[h]
  \centering
\includegraphics[width=0.46\textwidth,keepaspectratio,clip]{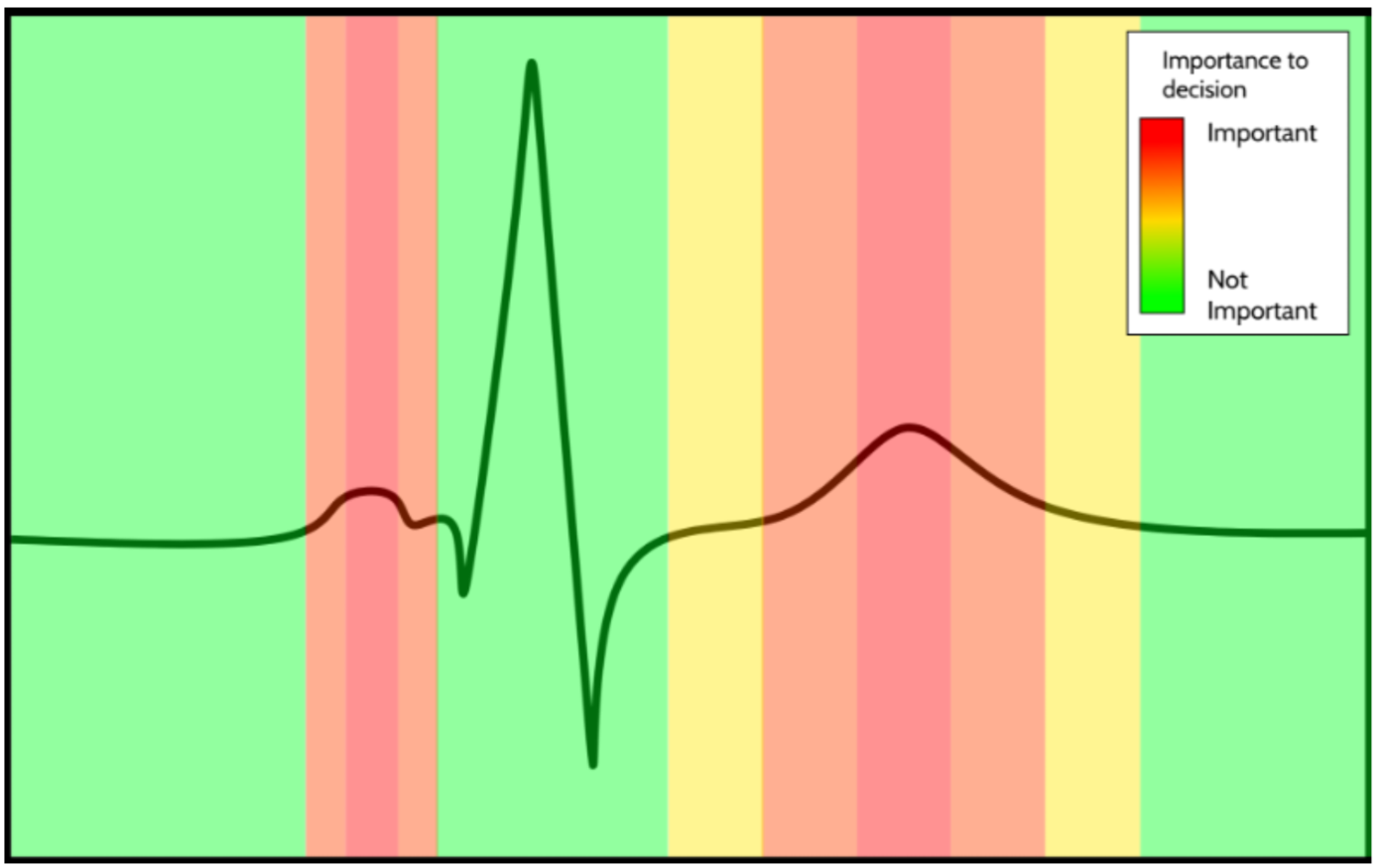}
  \caption{Vignette showing the saliency-map based explanations}
  \label{saliency-map}
\end{figure}

\begin{figure}[h]
  \centering
\includegraphics[width=0.46\textwidth,keepaspectratio,clip]{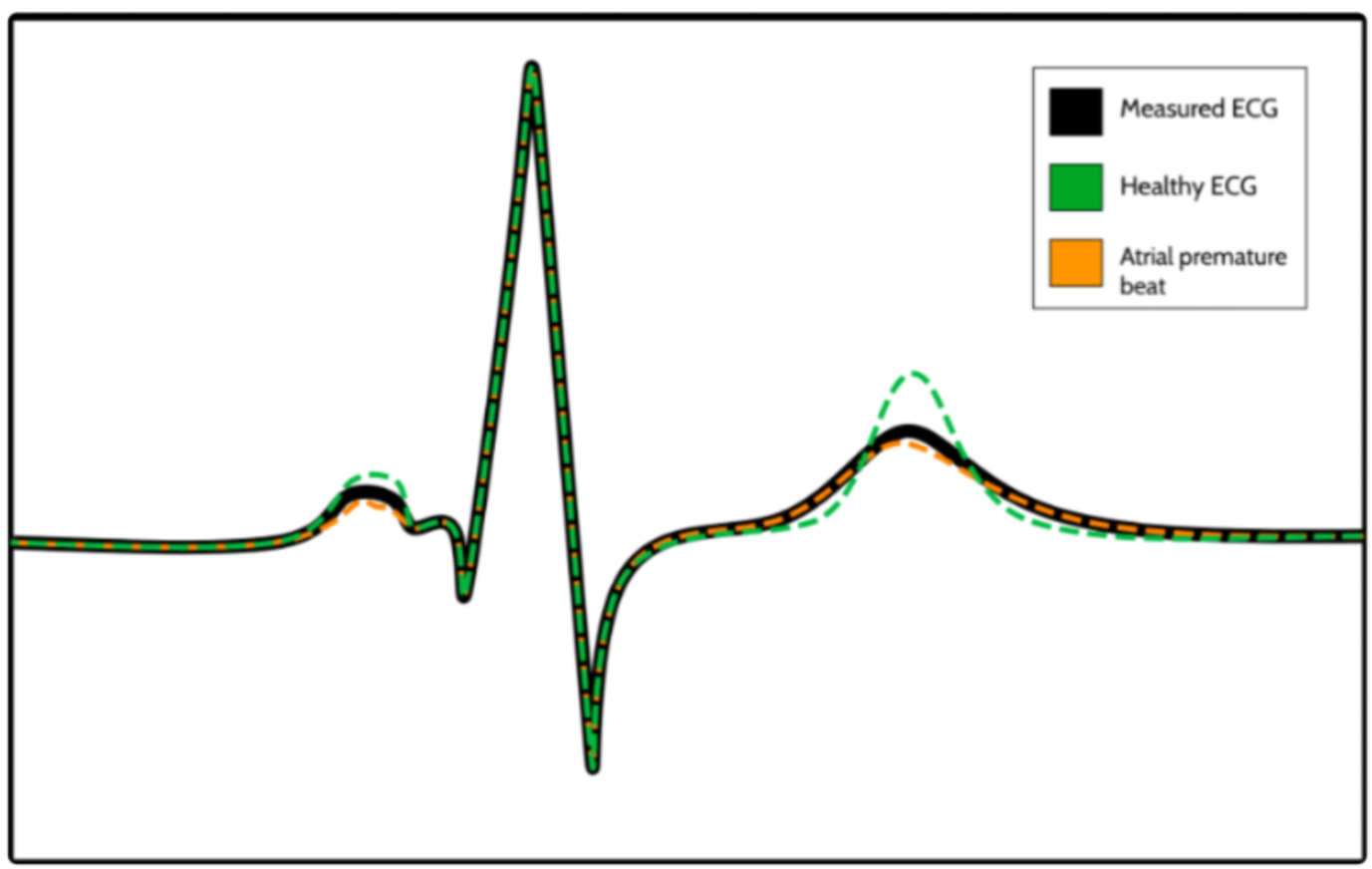}
  \caption{Vignette showing the counterfactuals-based explanation}
  \label{counterfactuals}
\end{figure}

To assess preferences regarding XAI approaches for arrhythmia detection, a survey was distributed to professionals in the medical field. The survey presented two different visualization strategies that an XAI method could provide. The first approach, a saliency map (Figure \ref{saliency-map}), highlights the relative importance of specific time segments in the ECG for decision-making using color coding, where red indicates segments contributing strongly to the predicted arrhythmia class and green indicates segments of low relevance. The second approach presents three signals: the measured ECG, a counterfactual normal signal, and an average signal corresponding to the predicted arrhythmia, as illustrated in Figure \ref{counterfactuals}. The counterfactual signal is generated by minimally modifying the original arrhythmic ECG so that it would be classified as normal, providing insight into the key features driving the model’s decision.

\subsection{XAI Methods}
The SHAP library includes several explainers that compute SHAP values using different underlying approaches. These approaches build upon fundamental XAI techniques discussed in Section \ref{rel_lit}, including: Permutation SHAP, which calculates SHAP values via random data permutations; Gradient SHAP, which leverages gradients similar to Grad-CAM \cite{Shrikumar_2019}; KernelSHAP, which uses a kernel-based approximation; and DeepSHAP, which is derived from DeepLIFT \cite{Shrikumar_2019}.

\section{Results}
\label{sec_results}
\subsection{Stakeholder input}
Input was collected from five clinical stakeholders, all of whom indicated a preference for saliency map visualizations, citing their effectiveness in clearly highlighting the regions of the ECG signal most relevant to the model’s decision-making process.

\subsection{Model Evaluation}
\begin{figure}[h]
  \centering
\includegraphics[width=0.46\textwidth,keepaspectratio,clip]{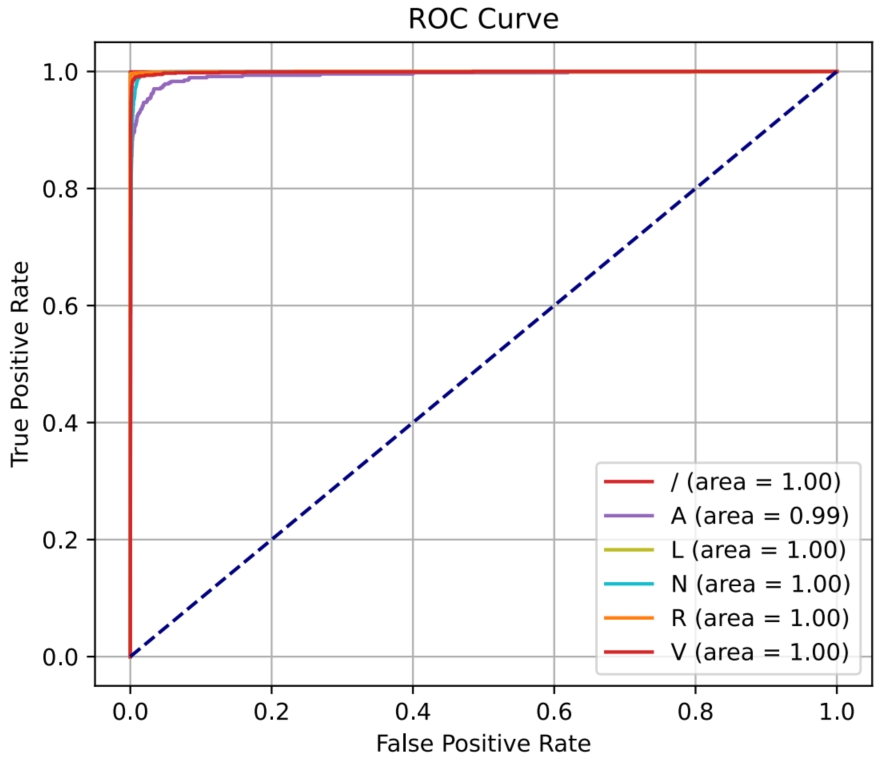}
  \caption{ROC curve of performance of the model on the validation set of the MIT-BIH dataset. In the legend, '/' indicates a paced beat, 'A' indicates an atrial premature beat, 'L' indicates a left bundle branch block, 'N' indicates a normal beat, 'R' indicates a right bundle branch block, and 'V' indicates a premature ventricular contraction.}
  \label{ROC}
\end{figure}

The model achieved a maximum training accuracy of 98.26\% and a validation accuracy of 98.30\% on the MIT-BIH database after 100 epochs. The classifier’s performance across the different arrhythmia classes is summarized in Table \ref{perf_arrhythmia}, and the ROC curve for the validation set is presented in Figure \ref{ROC}. These metrics were computed on the validation set.

\begin{figure}[h]
  \centering
\includegraphics[width=0.46\textwidth,keepaspectratio,clip]{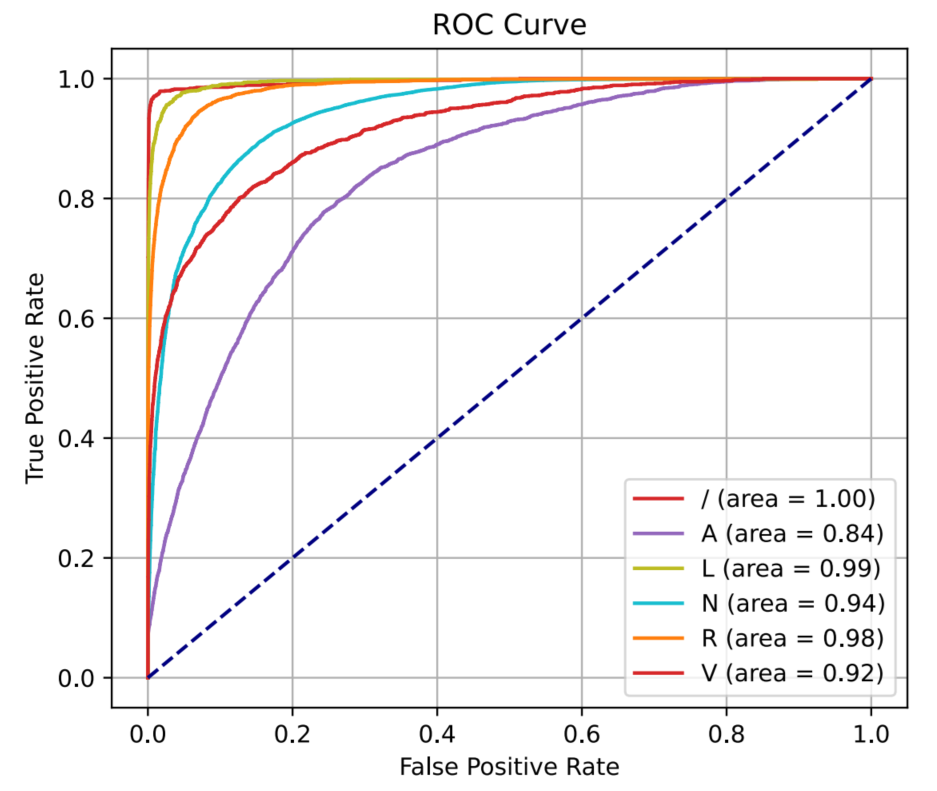}
  \caption{ROC curve of performance of the model on the validation set of the combined dataset. In the legend, '/' indicates a paced beat, 'A' indicates an atrial premature beat, 'L' indicates a left bundle branch block, 'N' indicates a normal beat, 'R' indicates a right bundle branch block, and 'V' indicates a premature ventricular contraction.}
  \label{ROC_combined}
\end{figure}

The removal of the LSTM layer resulted in a performance reduction of approximately 1–2\% compared to the results reported in \cite{Alamatsaz_2024}. When incorporating a second dataset, the model achieved a maximum training accuracy of 82.77\% and a validation accuracy of 73.45\%. The corresponding ROC curve is shown in Figure \ref{ROC_combined}, and detailed arrhythmia classification performance on the combined dataset is provided in Table \ref{perf_both}.

\begin{table}[]
    \centering
    \caption{Validation performance of the CNN-based arrhythmia detection model on the MIT-BIH dataset}
    \begin{tabular}{|c|c|c|c|c|}
\hline
  & Precision & Sensitivity & fl-score & Number of samples \\ \hline
Paced beat & 1.00 & 1.00 & 1.00 & 1420 \\ \hline
Atrial premature beat & 0.52 & 0.95 & 0.67 & 472 \\ \hline
Left bundle branch block & 0.99 & 0.99 & 0.99 & 1589 \\ \hline
Normal beat & 1.00 & 0.97 & 0.98 & 15070 \\ \hline
Right bundle branch block & 0.99 & 0.99 & 0.99 & 1474 \\ \hline
Premature ventricular contraction & 0.96 & 0.98 & 0.97 & 1387 \\ \hline
macro avg & 0.91 & 0.98 & 0.93 & 21412 \\ \hline
weighted avg & 0.98 & 0.97 & 0.98 & 21412 \\ \hline
accuracy & \multicolumn{3}{|c|}{0.97} & 21412 \\ \hline
\end{tabular}
\label{perf_arrhythmia}
\end{table}
\begin{table}[]
    \centering
    \caption{Validation performance of the CNN-based arrhythmia detection model on the combined dataset}
    \begin{tabular}{|c|c|c|c|c|}
\hline
  & Precision & Sensitivity & fl-score & Number of samples \\ \hline
Paced beat & 0.91 & 0.95 & 0.93 & 1483 \\ \hline
Atrial premature beat & 0.23 & 0.68 & 0.34 & 3224 \\ \hline
Left bundle branch block & 0.80 & 0.90 & 0.85 & 2077 \\ \hline
Normal beat & 0.97 & 0.71 & 0.82 & 33787 \\ \hline
Right bundle branch block & 0.75 & 0.80 & 0.77 & 2769 \\ \hline
Premature ventricular contraction & 0.33 & 0.70 & 0.45 & 1776 \\ \hline
macro avg & 0.67 & 0.79 & 0.69 & 45116 \\ \hline
weighted avg & 0.87 & 0.73 & 0.78 & 45116 \\ \hline
accuracy & \multicolumn{3}{|c|}{0.73} & 45116  \\ \hline
\end{tabular}
\label{perf_both}
\end{table}

\subsection{XAI Results}
\begin{figure*}[h]
  \centering
\includegraphics[width=0.99\textwidth,keepaspectratio,clip]{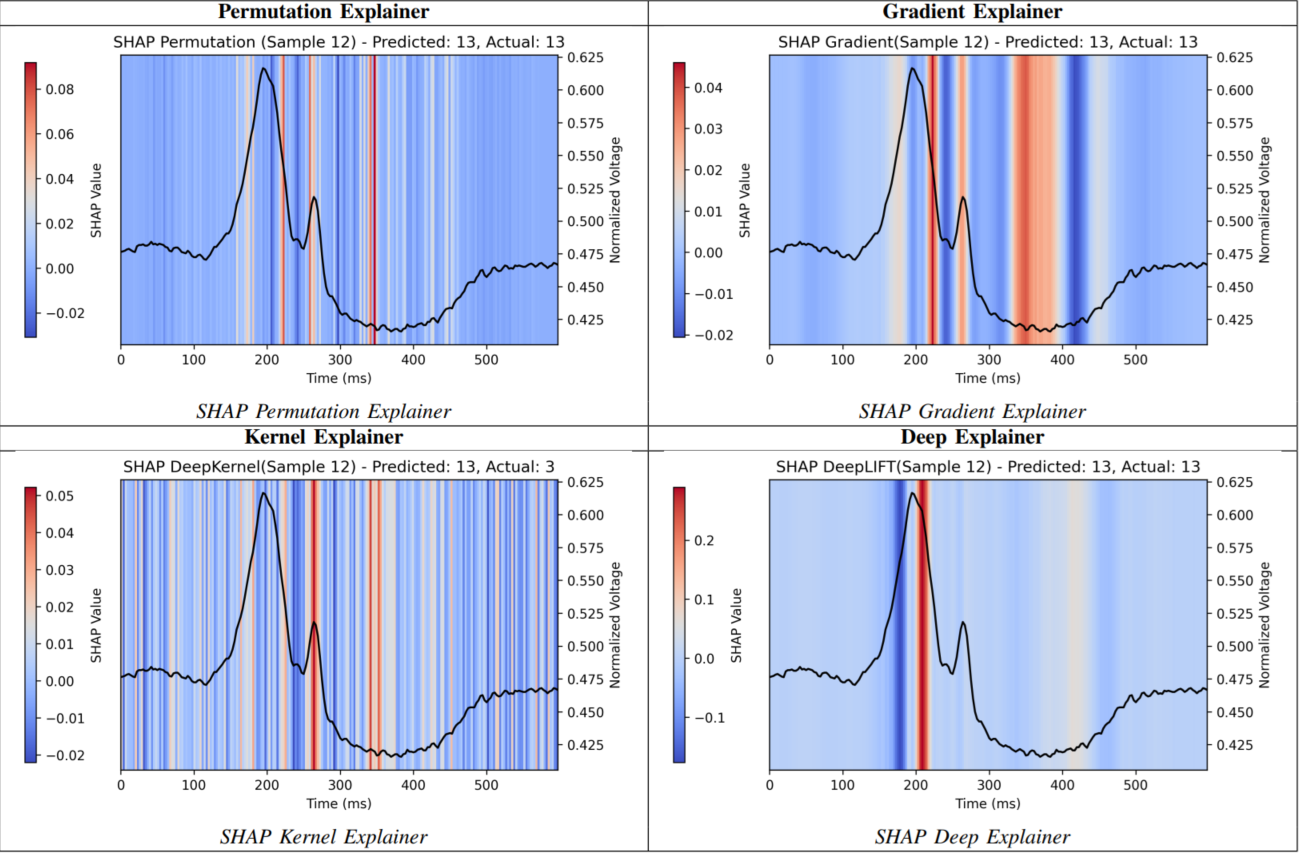}
  \caption{Comparison of the visualizations of different explainers}
  \label{explainer_comparison}
\end{figure*}

The results of the XAI analysis are presented as follows. A single heartbeat sample is illustrated using four different explainers—Permutation, Gradient, Kernel, and Deep—to demonstrate the corresponding visualizations (Figure \ref{explainer_comparison}). It can be observed that the Permutation and Kernel explainers provide fine-grained saliency maps but exhibit abrupt transitions, which may be confusing for users. In contrast, the Gradient and Deep explainers produce smoother and more interpretable visualizations, making them better suited for this application.

\begin{figure*}[h]
  \centering
\includegraphics[width=0.99\textwidth,keepaspectratio,clip]{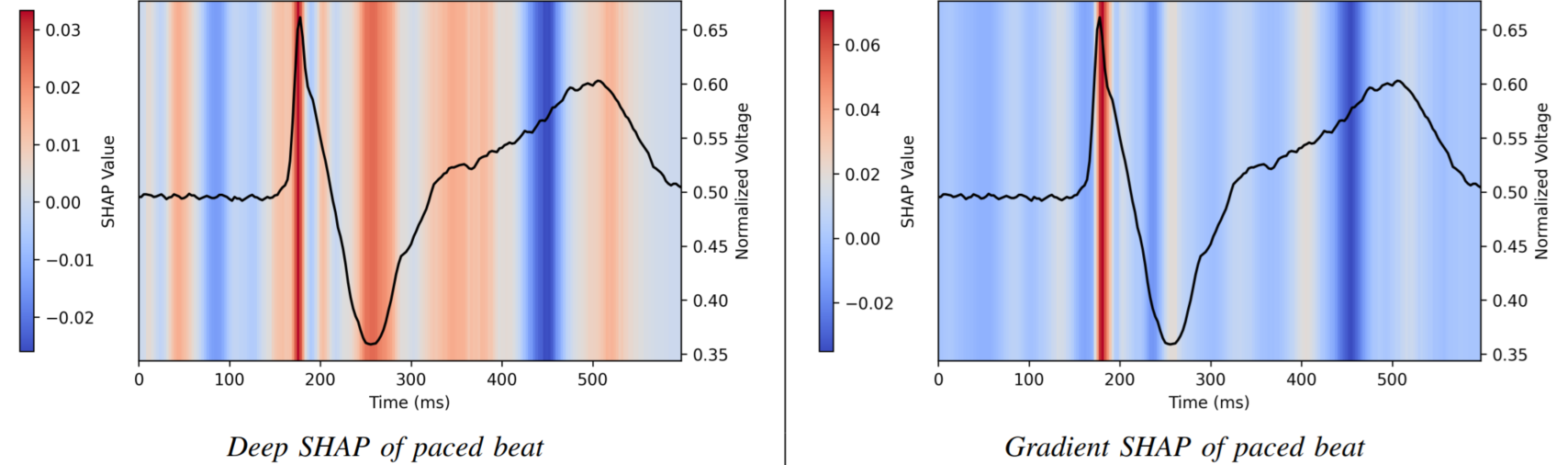}
  \caption{DeepLift and Gradient explainer applied to a paced beat when the model correctly labels it as a paced beat}
  \label{pb_comparison}
\end{figure*}

\begin{figure*}[h]
  \centering
\includegraphics[width=0.99\textwidth,keepaspectratio,clip]{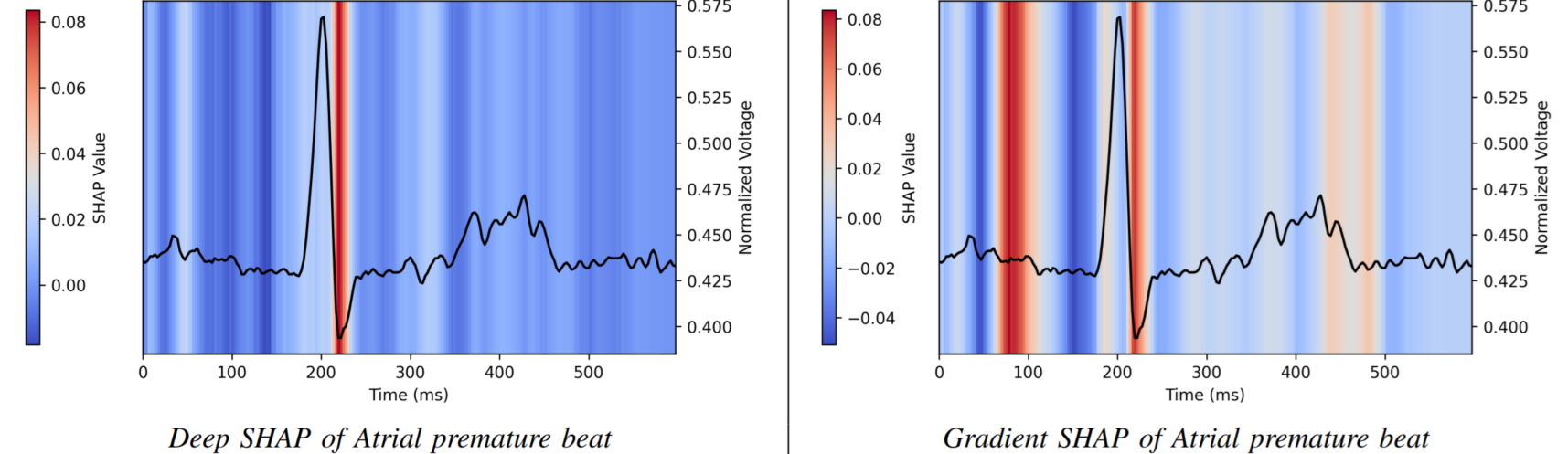}
  \caption{DeepLift and Gradient explainer applied to an atrial premature beat when the model correctly labels it as an atrial premature beat}
  \label{apb_comparison}
\end{figure*}

\begin{figure*}[h]
  \centering
\includegraphics[width=0.99\textwidth,keepaspectratio,clip]{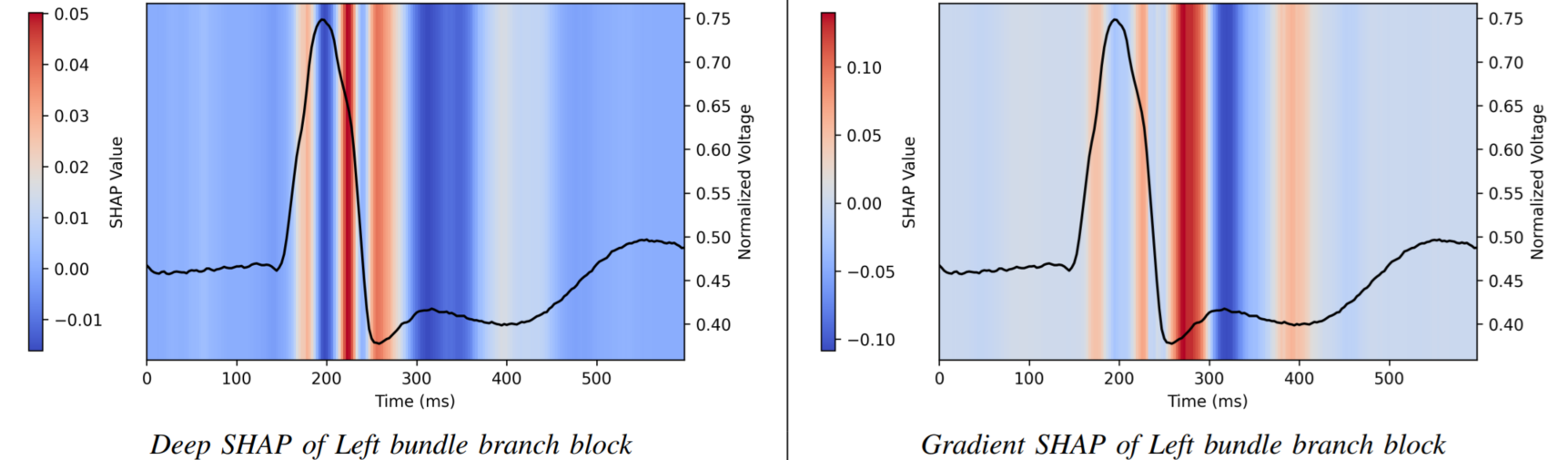}
  \caption{DeepLift and Gradient explainer applied to a left bundle branch block sample when the model correctly labels it as left bundle branch block}
  \label{lbbb_comparison}
\end{figure*}

\begin{figure*}[h]
  \centering
\includegraphics[width=0.99\textwidth,keepaspectratio,clip]{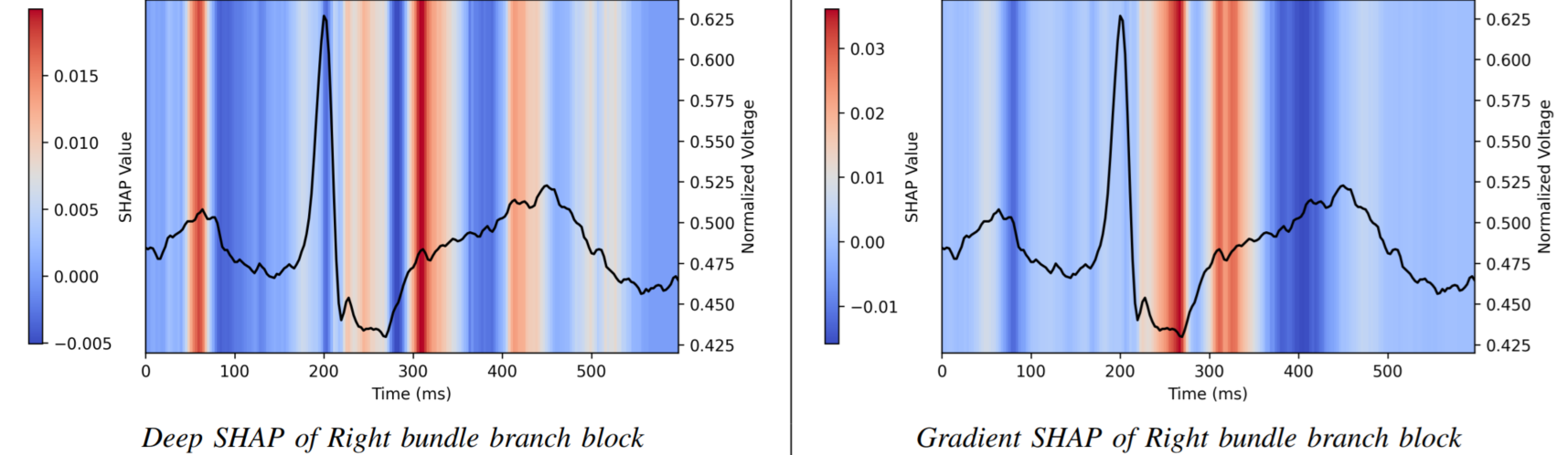}
  \caption{DeepLift and Gradient explainer applied to a right bundle branch block sample when the model correctly labels it as right bundle branch block}
  \label{rbbb_comparison}
\end{figure*}

\begin{figure*}[h]
  \centering
\includegraphics[width=0.99\textwidth,keepaspectratio,clip]{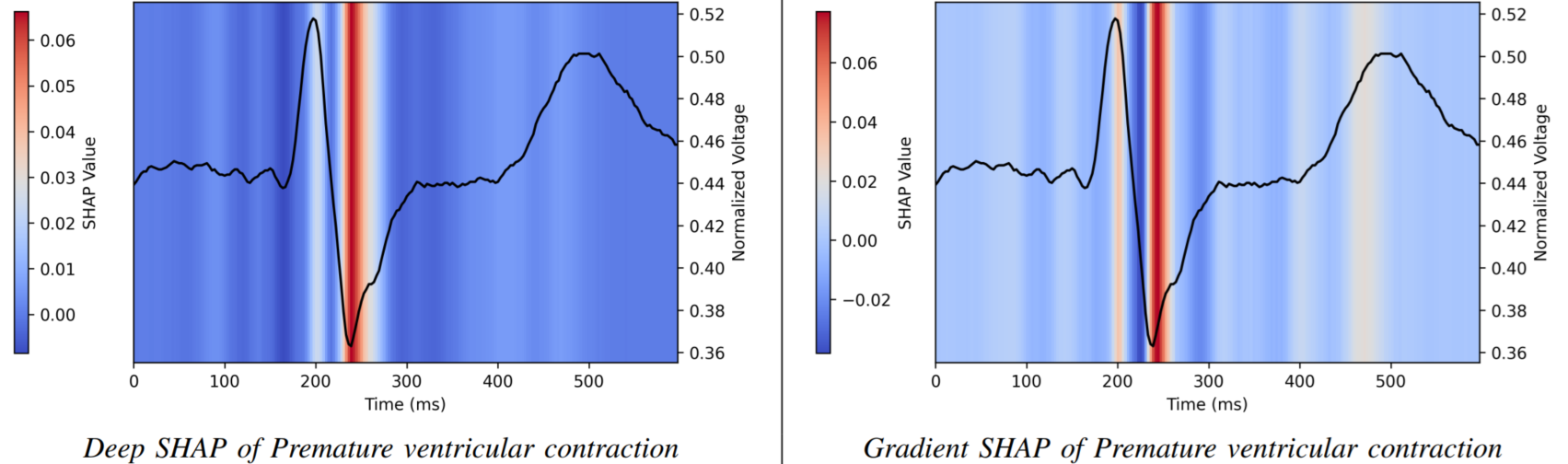}
  \caption{DeepLift and Gradient explainer applied to a right bundle branch block sample when the model correctly labels it as right bundle branch block}
  \label{PVC_comparison}
\end{figure*}

Accordingly, for each arrhythmia class, a representative sample is shown with both the Gradient and Deep explainers applied: Paced Beat (Figure \ref{pb_comparison}), Atrial Premature Beat (Figure \ref{apb_comparison}), Left Bundle Branch Block (Figure \ref{lbbb_comparison}), Right Bundle Branch Block (Figure \ref{rbbb_comparison}), and Premature Ventricular Contraction (Figure \ref{PVC_comparison}). While the explanations are not entirely consistent across all samples within a given arrhythmia class, even when the classifier correctly identifies them,they were selected to illustrate the general behavior of each XAI method for that class.

Figure \ref{pb_comparison} illustrates a paced beat, which is a heartbeat induced by a pacemaker and is characterized by a sharp peak caused by the device \cite{Hampton_2019}. The Gradient explainer emphasizes this peak more clearly than Deep SHAP, and both explainers highlight the peak’s importance as expected, providing an explanation that aligns with human understanding.

Figure \ref{apb_comparison} depicts an atrial premature beat, characterized by heartbeats originating in the atria that occur earlier than anticipated based on the underlying rhythm. In some cases, Gradient SHAP appears to capture residual signal from the preceding heartbeat, as these beats occur closer together than usual. However, this observation is not consistent across other samples, and the explainers for this arrhythmia do not consistently align with human understanding.

Figure \ref{lbbb_comparison} shows a left bundle branch block (LBBB), which is typically diagnosed by a widened QRS complex and specific QRS morphology changes in certain leads. Both explainers highlight regions corresponding to the widened QRS complex, indicating that the model’s decision-making aligns well with human understanding.

Figure \ref{rbbb_comparison} presents a right bundle branch block (RBBB), usually indicated by an M-shaped QRS complex or a depressed S wave. In the example shown, the signal exhibits a depressed S wave, which Gradient SHAP highlights more effectively than Deep SHAP, consistent with human understanding.

Figure \ref{PVC_comparison} depicts a premature ventricular contraction (PVC), which is generally diagnosed based on the absence or inversion of the P-wave. In this case, both explainers indicate that the model is prioritising regions of the signal corresponding to a trough that do not correspond to human diagnostic criteria (which could be caused by the model learning to priortise a negative P-wave in other signals). This suggests that, even when the model correctly classifies the arrhythmia, explainers can reveal inconsistencies or limitations in the model’s reasoning.

\section{Discussions}
\label{sec_discussions}
The results of training the model exclusively on the MIT-BIH dataset demonstrate a high overall accuracy. However, the atrial premature beat class represents an outlier, which can be logically explained. APBs are identified based on their occurrence earlier than expected within the cardiac rhythm \cite{Hampton_2019}. Since the model is trained on individual heartbeats, it lacks temporal context and cannot compare the interval between the current and preceding beat relative to the normal rhythm. This illustrates why static CNNs are insufficient, and motivates temporal fusion models that preserve inter-beat context.

The inclusion of the second dataset to the MIT-BIH dataset led to a notable decrease in model accuracy, with overall performance dropping by more than 20\%. This reduction is likely attributable to differences in measurement methodology between the 12-lead dataset and the MIT-BIH dataset. In particular, variations in lead configuration between the two datasets can produce substantially different signal characteristics, which may have introduced inconsistencies that affected the model’s performance. Differences in signal morphology across leads and devices require models that integrate information over multiple time windows, making predictions robust to irregularities and missing segments.

Feedback from clinical professionals indicated a clear preference for the saliency-map visualization (Figure \ref{saliency-map}) over the counterfactual approach (Figure \ref{counterfactuals}). However, several respondents noted that the counterfactual visualisation was not entirely clear, as the depiction of the arrhythmia did not accurately reflect the realistic waveform, which caused some confusion. One respondent further suggested that saliency maps would be more practical for multi-lead ECGs, as the counterfactual approach could become overly complex and visually chaotic when multiple signals are presented simultaneously.

Comparison of the four SHAP explainers revealed notable differences in interpretability. The Permutation explainer occasionally produced unreliable results due to random shuffling, reducing the consistency of the explanations. The Kernel explainer generally provided results similar to the Gradient explainer, although the output was less smooth. Both the Deep and Gradient explainers consistently highlighted clear points of interest, often converging on similar regions of the signal. Exceptions occurred where the two explainers disagreed on the most important segments. For paced beats, left bundle branch blocks, and right bundle branch blocks, the highlighted regions aligned well with human understanding of ECG-based aarhythmia diagnosis. In contrast, explanations for atrial premature beats did not align with human interpretation, likely reflecting the model’s lower classification performance on this class. For premature ventricular contractions, the explainers sometimes focused on the negative P-wave, whereas in other cases, attention was directed to deep negative deflections elsewhere in the signal. Post-hoc saliency maps cannot capture higher-level temporal dependencies, reinforcing the need for intrinsic, memory-based explainability embedded in temporal fusion architectures. It should be emphasized that, in the absence of validation from trained medical professionals, connections between these XAI results and established medical knowledge should be considered suggestive rather than definitive.

\section{Conclusions}
\label{sec_conclusions}
This study evaluated multiple explainability methods for CNN-based arrhythmia detection and highlighted several critical challenges. We found that saliency methods such as GradCAM and DeepLIFT were more effective than simpler perturbation-based methods, as they highlighted waveform regions relevant to expert interpretation. However, even the stronger methods failed for arrhythmias like atrial premature beats (APBs), where diagnostic context depends on the timing and relationship between successive beats. This limitation reflects a broader gap: static CNNs that analyze isolated segments cannot capture temporal dependencies essential for clinical reasoning. Performance differences across datasets further underscored this challenge. Variability in signal morphology across leads and acquisition settings revealed the need for models capable of integrating information across multiple time windows, making predictions robust to irregular sampling, missing data, and heterogeneous inputs. Moreover, post-hoc explanation methods, while useful, do not intrinsically incorporate memory or contextual reasoning, limiting their ability to provide clinically faithful explanations. Our findings therefore demonstrate that reliable explainability in arrhythmia detection—and in healthcare time-series more generally—cannot be achieved without temporal models that explicitly integrate past and present data. Temporal fusion architectures, especially those embedding attention-based mechanisms, offer a methodological pathway to capture features-of-change, ensure robustness across variable datasets, and provide human-aligned, context-aware explanations. Developing such architectures represents the next step toward clinically meaningful, explainable AI for decision support.

\bibliographystyle{unsrt}  
\bibliography{references}  


\end{document}